\documentclass[journal]{IEEEtran}

\usepackage{multirow}
\usepackage[normalem]{ulem}
\useunder{\uline}{\ul}{}
\usepackage{amsthm,amsmath,amssymb}
\usepackage{mathrsfs}
\usepackage{times}
\usepackage{mathrsfs}
\usepackage{epsfig}
\usepackage{graphicx}
\usepackage{amsmath}
\usepackage{amssymb}
\usepackage{graphicx}
\usepackage{cite}
\usepackage{enumerate}
\usepackage{cases}
\usepackage{multirow}
\usepackage{verbatim}
\usepackage{amssymb}
\usepackage{CJK}
\usepackage{algorithm}
\usepackage{algorithmicx}
\usepackage{algpseudocode}
\usepackage{color}
\usepackage{bm}
\usepackage{booktabs}
\usepackage[colorlinks,linkcolor=red]{hyperref}
\usepackage{amssymb}
\usepackage{hyperref}
\usepackage{marvosym}
\usepackage{subfigure}
\usepackage{color}

\newcommand{\etal}{\textit{et al}.}
\newcommand{\ie}{\textit{i}.\textit{e}.}

\newcommand{\etc}{\textit{etc}}

\begin{document}

\title{Does Thermal Really Always Matter for RGB-T Salient Object Detection?}

\author
{
 Runmin Cong,~\IEEEmembership{Member,~IEEE,} Kepu Zhang,  Chen Zhang, Feng Zheng, Yao Zhao,~\IEEEmembership{Senior Member,~IEEE,}\\ Qingming Huang,~\IEEEmembership{Fellow,~IEEE,} and Sam Kwong,~\IEEEmembership{Fellow,~IEEE}

\thanks{Runmin Cong is with the Institute of Information Science, Beijing Jiaotong University, Beijing 100044, China, also with the Beijing Key Laboratory of Advanced Information Science and Network Technology, Beijing 100044, China, and also with the Department of Computer Science, City University of Hong Kong, Hong Kong SAR, China (e-mail: rmcong@bjtu.edu.cn).}
\thanks{Kepu Zhang, Chen Zhang, and Yao Zhao are with the Institute of Information Science, Beijing Jiaotong University, Beijing 100044, China, and also with the Beijing Key Laboratory of Advanced Information Science and Network Technology, Beijing 100044, China (e-mail: kpzhang@bjtu.edu.cn; chen.zhang@bjtu.edu.cn; yzhao@bjtu.edu.cn).}
\thanks{Feng Zheng is with the Department of Computer Science and Technology, Southern University of Science and Technology, Shenzhen 518055, China, and also with the Research Institute of Trustworthy Autonomous Systems, Shenzhen 518055, China (e-mail: f.zheng@ieee.org).}
\thanks{Qingming Huang is with the School of Computer Science and Technology, University of Chinese Academy of Sciences, Beijing 101408, China, also with the Key Laboratory of Intelligent Information Processing, Institute of Computing Technology, Chinese Academy of Sciences, Beijing 100190, China, and also with Peng Cheng Laboratory, Shenzhen 518055, China
(email: qmhuang@ucas.ac.cn).}
\thanks{Sam Kwong is with the Department of Computer Science, City University of Hong Kong, Hong Kong SAR, China, and also with the City University of Hong Kong Shenzhen Research Institute, Shenzhen 51800, China (e-mail: cssamk@cityu.edu.hk).}

}

\markboth{}
{Shell \MakeLowercase{\textit{\etal}}: Bare Demo of IEEEtran.cls for IEEE Journals}
\maketitle

\begin{abstract}
In recent years, RGB-T salient object detection (SOD) has attracted continuous attention, which makes it possible to identify salient objects in environments such as low light by introducing thermal image. However, most of the existing RGB-T SOD models focus on how to perform cross-modality feature fusion, ignoring whether thermal image is really always matter in SOD task. Starting from the definition and nature of this task, this paper rethinks the connotation of thermal modality, and proposes a network named TNet to solve the RGB-T SOD task. In this paper, we introduce a global illumination estimation module to predict the global illuminance score of the image, so as to regulate the role played by the two modalities. In addition, considering the role of thermal modality, we set up different cross-modality interaction mechanisms in the encoding phase and the decoding phase. On the one hand, we introduce a semantic constraint provider to enrich the semantics of thermal images in the encoding phase, which makes thermal modality more suitable for the SOD task. On the other hand, we introduce a two-stage localization and complementation module in the decoding phase to transfer object localization cue and internal integrity cue in thermal features to the RGB modality.
Extensive experiments on three datasets show that the proposed TNet achieves competitive performance compared with 20 state-of-the-art methods. The code and results can be found from the link of \url{https://rmcong.github.io/proj\_TNet.html}.
\end{abstract}


\IEEEpeerreviewmaketitle

\section{Introduction} \label{sec1}

\begin{figure}[!t]
\centering
\centerline{\includegraphics[width=9cm,height=5cm]{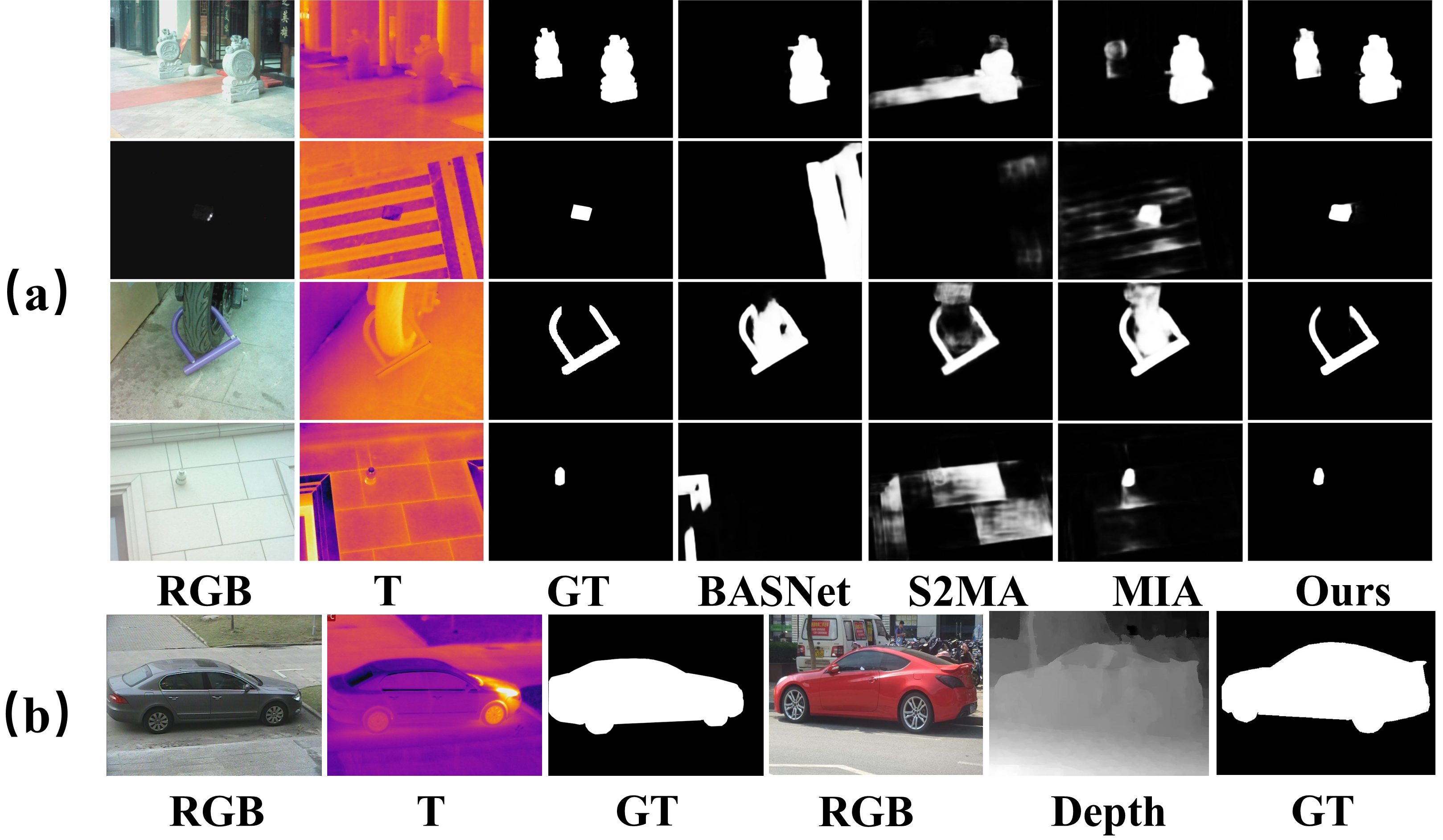}}
\caption{(a) Some visual examples of different SOD methods, including the RGB SOD (\ie, BASNet \cite{qin2019basnet}), RGB-D SOD (\ie, S2MA \cite{liu2020learning}), RGB-T SOD (\ie, MIA \cite{liang2022multi} and Ours). (b) Comparison of RGB-D and RGB-T data in similar scenarios for the SOD task.}
\label{fig1}
\end{figure}

\IEEEPARstart{S}{ALIENT} object detection (SOD) aims to locate the objects in an image that most attract the human visual attention, which has been widely used in many related fields \cite{crm/tcsvt19/review}, such as object segmentation \cite{wang2017saliency,crm/ACMMM20/DMVOS,crm/ins21/superpixel}, object detection\cite{9424966}, visual tracking \cite{lu2021rgbt}, content enhancement \cite{crm/JEI16/underwater,crm/tip21/underwaterMedium,crm/spl21/underwater,crm/cvpr20/low-light,crm/tmm20/dehazing,crm/tits22/low-light,crm/tmm22/blindSR,crm/acmmm21/bridgenet,crm/CVPR21/depthSR,crm/tip19/depthSR,crm/ijcai20/SR,crm/mtap22/dehazing}, and quality assessment \cite{crm/SPIC21/underwaterIQA,crm/tcsv22/underwaterIQA}. 
In recent years, deep learning technology has driven SOD task to achieve amazing performance with its powerful feature representation capabilities under ideal circumstances\cite{hou2017deeply,ma2019salient,liu2018picanet,han2014background,zhang2019synthesizing,li2016visual,9142398,9505635,wang2021reconcile,zhu2019pdnet,wu2019cascaded,zhao2019egnet,qin2019basnet,tmm2,crm/aaai20/GCPANet,crm/tcyb22/rsi,crm/nc20/rsi,crm/tip21/DAFNet}. But when encountering some challenging scenes such as low light or darkness, the performance of this single-modality SOD will degrade significantly. For example, as shown in the second row of Fig. \ref{fig1}(a), the extremely low-light scenes make it difficult for humans to clearly distinguish the salient object, and the state-of-the-art BASNet \cite{qin2019basnet} (an RGB SOD method) also fails to detect the salient object at all. To address this challenging scene, we can introduce infrared thermal images for auxiliary discrimination from the perspective of physical equipment. The infrared thermal imaging sensor can capture the infrared radiation emitted by the object. The higher the temperature of the object, the stronger the infrared radiation. The sensor collects different heat differences, and the thermal image is processed by electronic technology, which reflects the temperature distribution on the surface of the object. The second column of Fig. \ref{fig1}(a) provides some examples of thermal images that are very intuitive and effective complements to the RGB images in the low-light scenes. In this way, with the introduction of thermal modality, a new SOD branch called RGB-T SOD was born. 

Since its inception, RGB-T SOD task is compared with its sibling RGB-D SOD task, where `D' means depth map.
Some visual examples are shown in Fig. \ref{fig1}(b). Upon observation, the fundamental difference between the two tasks is the physical meaning of the additional modality. For the RGB-D SOD task, the depth information of the scene can be directly perceived by humans, which describes the distance relationship of objects and directly affects people's recognition of salient objects. In general, distant and small objects are not salient to the human eye. However, the thermal image cannot be directly observed by the human eyes without some special equipment, which mainly reflects the temperature of the object and is not directly related to the identification of salient objects. In fact, there is no prior assumption that the hotter the target, the more salient it is. It is even possible that the salient objects in the RGB image and thermal image are inconsistent. For example, in the first row of Fig. \ref{fig1}(a), the salient objects in the RGB image are two stone piers, while the objects with higher temperature in the thermal image are the pillars in the background.  
At this time, the saliency ground-truth annotations of samples are also mainly based on the RGB image rather than the thermal image. This is clearly different from the RGB-D SOD task, and thus leads to suboptimal results on the RGB-T task by simply porting the RGB-D SOD model directly. For example, the fifth column in Fig. \ref{fig1}(a) is the result of using RGB-D SOD method (\ie,  S2MA \cite{liu2020learning}) for the RGB-T SOD task, resulting in poor predictions. In this way, the timing and strategy of thermal image introduction is the key to the RGB-T SOD task.

Let's start with the first question, when to use the thermal modality information? As mentioned earlier, the introduction of thermal images in SOD task is mainly used to address challenging scenes, such as low light. If the cross-modality interaction is performed indiscriminately like the RGB-D SOD task, it does not actually solve the problem of the RGB-T SOD task itself, and may even lead to performance degradation. Therefore, starting from the nature and definition of the task, choosing the right time to apply the thermal image is critical and luxurious. In other words, it generally should play a more important role in low-light and dark scenes. Based on this, we introduce a Global Illumination Estimation (GIE) module to predict the global illuminance score of an image, so as to control and regulate the interaction between the RGB image and the thermal image. Specifically, we first introduce a pretrained deep Retinex decomposition network \cite{zhang2021beyond} to generate the corresponding illumination map from the input image, which describes the brightness intensity information of scene regions. Then, we quantize it into an illuminance score through global average pooling and sigmoid activation, which is used to control how much of the thermal modality is introduced in the RGB-T SOD model. 

The second twice-told story is the cross-modality interaction problem. Rethinking the connotation of the thermal image in the SOD task, it is actually difficult for people to intuitively perceive the thermal image, so it does not have clear and explicit semantic information like the depth map. Therefore, in order to make the thermal image more suitable for the SOD task, we introduce a Semantic Constraint Provider (SCP) module in the encoding stage to enrich the semantic attributes for each layer of the thermal image branch. In this way, we can bridge the gap between the thermal image and SOD task, thereby providing more reliable information for cross-modality interaction and saliency decoding. 
In the decoding stage, we re-examine the functions of the two modalities and design a Localization and Complementation (LC) module, where the thermal features play an auxiliary role for the RGB features. On the one hand, considering that the thermal image plays a certain role in determining the location of salient objects, especially in low-light and dark scenes. Therefore, in each layer of decoding, we use the thermal features to generate a spatial mask to assist in the object localization of the RGB image, with the constraint of global illuminance score. On the other hand, due to the limitation of the thermal modality itself, it may be difficult to achieve good results with thermal supplementation directly. Therefore, we adaptively implement cross-modality information interaction using global illuminance score, and use it as the encoding skip connection features to supplement the localization-corrected RGB decoding features.

The main contributions of this paper can be summarized as follows:
\begin{itemize}
\item We rethink the value and role of thermal images in SOD task, and propose a global illumination estimation module to control and regulate the interaction between the RGB image and the thermal image, thereby better adapting to challenging scenes such as low light.
\item In the encoding stage, a semantic constraint provider is designed to supplement the semantic content for each layer of the thermal image branch, which makes the thermal features more suitable for the SOD task.
\item In the decoding stage, a localization and complementation module is developed, which uses thermal features to provide effective object localization and integrity information for RGB decoding features. 
\item The proposed network achieves superior performance compared to 20 state-of-the-art methods on three public benchmark datasets.
\end{itemize}

The rest of this paper is organized as follows: Section \ref{sec2} presents related work of SOD, then Section \ref{sec3} details our proposed RGB-T SOD model. Next, Section \ref{sec4} presents our experimental results and corresponding analyses. Finally, Section \ref{sec5} is our conclusion.

\begin{figure*}[!t]
\centering
\centerline{\includegraphics[width=1\linewidth]{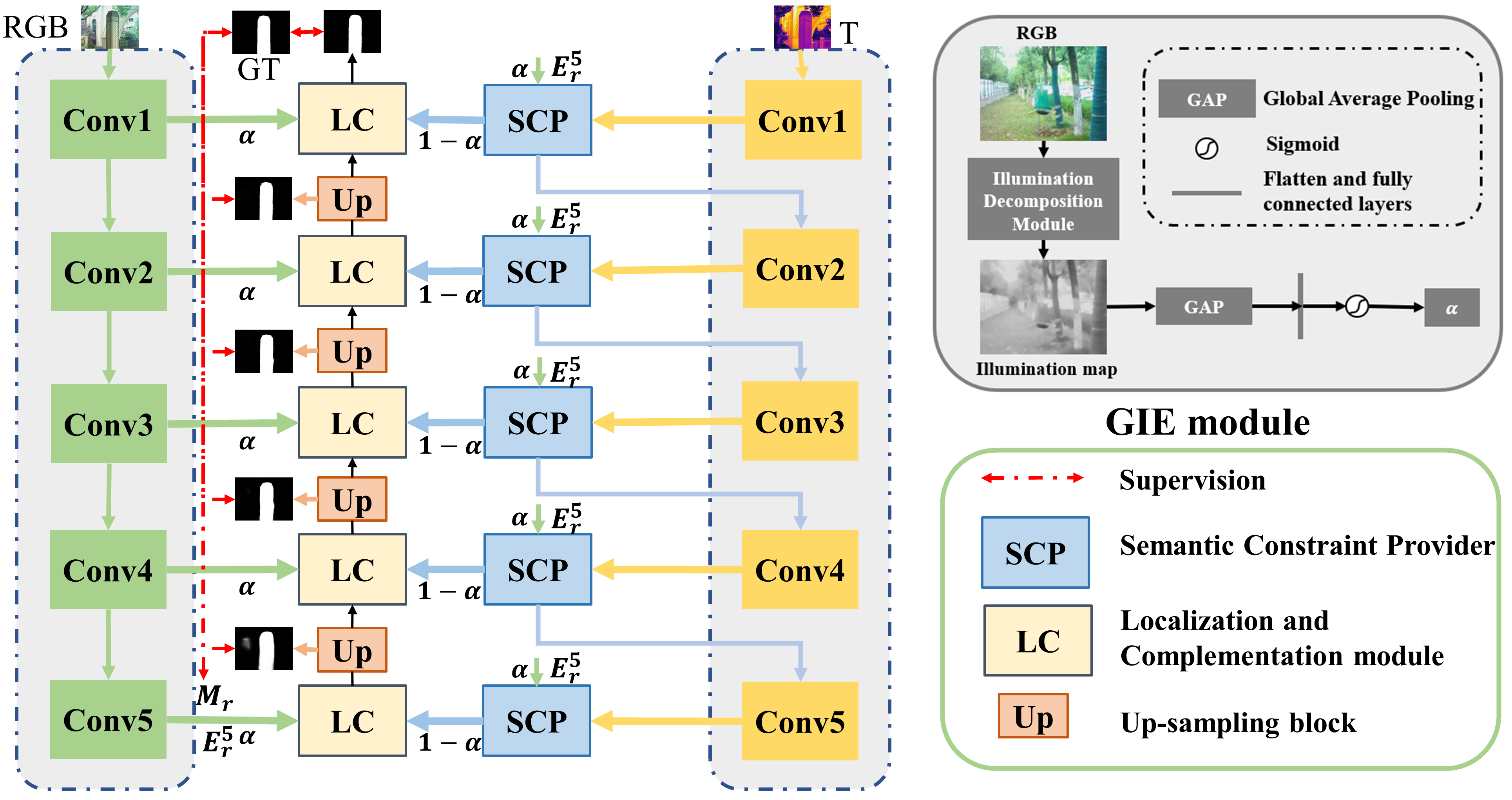}}
\caption{The overview architecture of the proposed TNet, which follows an encoder-decoder structure. The upper right corner is the global illumination estimation (GIE) module, predicting a global illuminance score to control the role of the two modalities. In the dual-stream encoding, the semantic constraint provider (SCP) module is introduced to enrich the thermal features with the help of high-level RGB semantics, making it more suitable for SOD tasks. 
Finally, the single-stream decoding with the RGB modality as the dominant and the thermal modality as the auxiliary implements layer-by-layer decoding under the action of the localization and complementation (LC) module, thereby obtaining the final saliency prediction.
}
\label{fig2}
\end{figure*}

\section{Related Work} \label{sec2}
\subsection{RGB and RGB-D Salient Object Detection}
In recent years, with the rapid development of deep learning, RGB SOD \cite{wu2019cascaded,zhao2019egnet,qin2019basnet,crm/aaai20/GCPANet,tmm2,crm/tcyb22/rsi,crm/nc20/rsi,crm/tip21/DAFNet} and RGB-D SOD \cite{piao2019depth,crm/tcyb21/ASIFNet,fu2020jl,crm/tip21/DPANet,crm/tip21/DynamicRGBDSOD,crm/eccv20/RGBDSOD,fdp1,crm/acmmm21/CDINet,fdp2,fdp3,fdp4,tmm1,tmm3,tmm4,8704255,6505832,9406382,zhang2021depth,chen2021rgb,crm/tmm22/3DSaliency,crm/tip22/CIRNet} methods achieved ground-breaking performance.
Qin \etal \cite{qin2019basnet} focused on better boundary quality, introduced the densely supervised network and residual optimization modules, and designed a new hybrid loss for boundary-aware salient object detection. Wu \etal\cite{wu2019cascaded} extracted high-level features through the attention branch and detection branch, achieved a better saliency map through HAM module connection and improved running speed by discarding low-level features. Zhao \etal \cite{zhao2019egnet} introduced the edge information into the SOD task. 

As effective auxiliary information, the depth map has been introduced into the SOD model to address some challenging and complex scenes. 
For example, Piao \etal \cite{piao2019depth} proposed a depth-sensing multi-scale weighting module to explore the relationship between depth information and multi-scale objects and designed a recurrent attention module to iteratively generate better saliency results. In \cite{liu2020learning}, a new intermediate fusion strategy was proposed to accurately locate the salient object. Fu \etal \cite{fu2020jl} introduced a joint learning and dense collaboration module to solve the RGB-D salient object detection task based on Siamese network. Considering the different roles of the RGB and depth modalities, Zhang \etal \cite{crm/acmmm21/CDINet} proposed a cross-modal differential interaction mode for RGB-D SOD task, so that the RGB and depth branches can play their respective advantages and complement each other better, resulting in better saliency results.

Although the performance of RGB SOD and RGB-D SOD methods is promising, they still struggle to handle the dark or low-light scenes. Therefore, the RGB-T SOD task that introduced thermal images emerged and developed rapidly.

\subsection{RGB-T Salient Object Detection}

Most of the traditional RGB-T SOD methods are based on graph-related technologies. Wang \etal \cite{wang2018rgb} presented the first relevant RGB-T SOD dataset, named VT821, and proposed a multi-task manifold sorting algorithm to solve the task. 
Tu \etal \cite{tu2019m3s} also used manifold sorting to achieve multi-modal and multi-scale fusion of different features and then introduced intermediate variables to infer the optimal sorting seeds in manifold sorting.
In addition, Tu \etal \cite{tu2019rgb} proposed a collaborative graph learning method to solve the RGB-T SOD task. However, with the rapid development of deep learning, compared to the performance of CNN-based RGB-T SOD methods, traditional methods are far behind.
For the CNN-based RGB-T SOD methods, how to extract single-modality and integrate cross-modality cues is the most important issue. Zhang \etal \cite{tu2021multi} proposed a multi-interaction dual decoder to better fuse cross-modality features, multi-level features and global contextual features.  Gao \etal \cite{9439490} performed multiple stages and multiple scales feature fusion in RGB-T SOD. Wang \etal \cite{9493207} proposed a novel cross-guided fusion network to perform adequate cross-modality fusion and took full advantage of high-level semantic information. Zhou \etal \cite{9420662} used an efficient bilateral fusion, multi-level coherent fusion module for cross-modality fusion. Zhou \etal \cite{9583676} proposed a perceptual importance fusion module for the fusion of different modalities. Liang \etal \cite{liang2022multi} proposed a multi-modality interactive attention unit to capture the single-modality multi-layer contextual features, and two decoding modules to achieve multi-source and multi-level feature fusion.

Although many of the above works have achieved competitive performance, they do not clearly define the relationship between thermal image and saliency attributes. Therefore, we rethink the value and role of thermal images in SOD task, and propose a new RGB-T SOD network, named TNet, which regulates the interaction between the RGB image and the thermal image via the GIE module, and achieves the cross-modality interaction via the SCP module in the feature encoding and the LC module in the feature decoding. 


\section{Proposed Method} \label{sec3}

\subsection{Architecture Overview}
In this paper, our main idea is to find the right time to introduce the thermal information into the SOD model by rethinking the connotation and function of thermal modality. Therefore, we proposed a two-stream encoder-decoder network to achieve RGB-T SOD, named TNet, as shown in Fig. \ref{fig2}. Before we start officially, the input RGB image is first embedded into a Global Illumination Estimation (GIE) module to predict a global illuminance score $\alpha$, which is used to control and regulate the interaction between the RGB image and the thermal image in the encoder and decoder stages. 
In the feature decoder, we use the ResNet50 \cite{he2016deep} that removes the last pooling layer and fully connected layer as the backbone to extract the multi-level encoder features of RGB image and thermal image, denoted as $E_r^i$ and $E_t^i$ $(i=\{1,2,…,5\})$, respectively. Then, in order to supplement the semantics for the thermal features, the top-level RGB features containing rich semantic information are input into the Semantic Constraint Provider (SCP) module, so that the thermal features are semantically guided and refined under the constraint of the global illuminance score $\alpha$. 
In the decoding stage, we set up a single-stream decoder with the RGB modality as the dominant features, and the corresponding RGB encoder features and semantic-embedded thermal features are sent to the Localization and Complementation (LC) module for step-by-step decoding. The LC module is also constrained by the global illuminance score $\alpha$, and the output of the last LC module is the final predicted saliency map. In the following subsections, we will introduce the details of GIE module, SCP module, and LC module one by one.

\subsection{Global Illumination Estimation (GIE) module}
According to the definition and setting of salient objects in RGB-T images, people still mainly rely on RGB images to determine whether objects are salient or not, while thermal image is more to provide some auxiliary information to deal with some challenging scenes, such as low light. In addition, as discussed in Section \ref{sec1}, if the network relies too much on the thermal image, it may lead to incomplete structures and even conflict with the real salient object. Therefore, in the proposed TNet, we hope that thermal image can play a more important role in low-light scenes, and design a global illumination estimation (GIE) module to describe the brightness information of the scene, as shown in the upper right of Fig. \ref{fig2}. The estimated illuminance score can be further used to regulate the role of the thermal image in the SOD model and guide the interaction between the RGB and thermal modalities.

Inspired by the low-light enhancement method based on Retinex theory, the image $I$ can be decomposed into the reflectivity component $R$ and illuminance component $L$, that is, $I=R\otimes L$, where $\otimes$ denotes the element-wise multiplication. The illuminance map reflects the brightness distribution of the image, which can be used to distinguish whether the scene is low-light or normal-light. Inspired by this, a pretrained Retinex decomposition network \cite{zhang2021beyond} is introduced to obtain the illuminance map, which can be formulated as:
\begin{equation}\label{illuma}
G_L=\mathcal{F} (I,\theta)
\end{equation}
where $G_L$ is the illuminance map, $I$ is the input RGB image, and $\mathcal{F}$ denotes the pretrained Retinex decomposition network \cite{zhang2021beyond}, and $\theta$ is the learnable parameters. 

Our original intention is to obtain an illuminance measurement to describe the brightness information of the scene, which is further used to regulate the role of the thermal image in the SOD model and guide the interaction between the RGB and thermal modalities. Moreover, for the obtained illuminance measurement reflecting the light distribution of an image, we only need to get a quantitative value that describes the light intensity of the entire scene, and do not need to obtain a pixel-by-pixel light map. In addition, the pixel-by-pixel light map may bring noise interference, since there is no prior assumption that brighter areas are more salient. Based on this, we quantize the illuminance map into an illuminance score through the global average pooling and sigmoid activation:
\begin{equation}\label{illuma}
\alpha =\delta (FC(flatten(GAP(G_L))))
\end{equation}
where $\delta (\cdot )$ is the sigmoid activation function, $flaten (\cdot )$ is a flatten layer that changes the dimension from $4$ to $2$, $FC (\cdot )$ denotes the fully connected layer, and $GAP (\cdot )$ is the global average pooling layer. 

The larger the global illuminance score $\alpha$, the higher the brightness of the scene, and it will be used in the entire encoding and decoding stages to control the introduction of thermal information. In the encoding process, we use it to control how much the semantic constraints of the RGB features affect the thermal features. When the scene is darker with a smaller global illuminance score, the degree of RGB semantic provider to thermal features will be reduced, thereby preserving the original thermal features as much as possible. In the decoding process, we use it to adjust the influence of the thermal features on localization and complementation of RGB features. The darker the scene, the greater the role of the thermal features. More details will be introduced in the following Sections \ref{sec3}-C and \ref{sec3}-D.

\subsection{Semantic Constraint Provider (SCP) module}
\begin{figure}[!t]
\centering
\centerline{\includegraphics[width=1\linewidth]{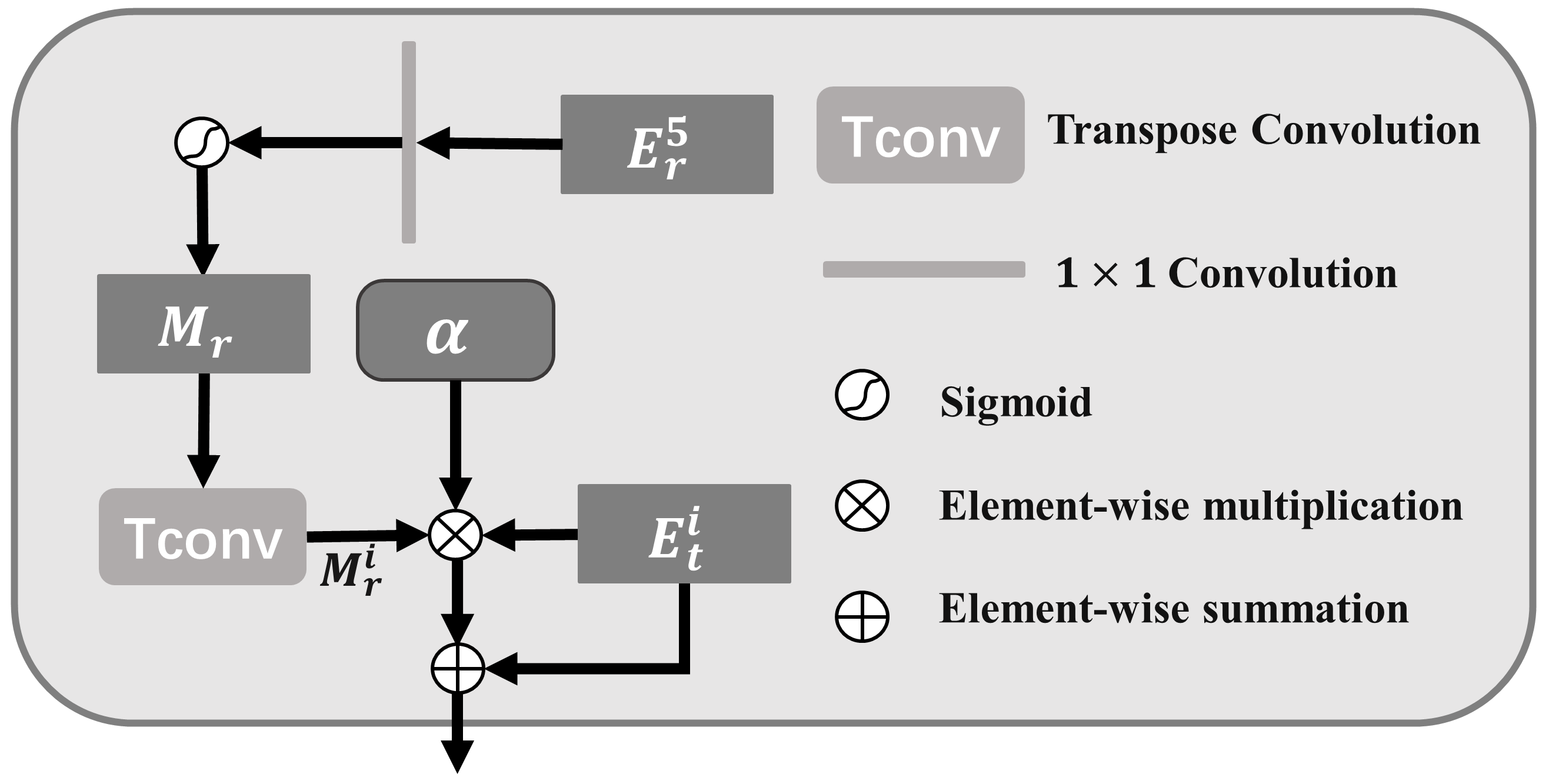}}
\caption{Structure diagram of the Semantic Constraint Provider.}
\label{fig3}
\end{figure}

As mentioned earlier, there is always a weak correlation between thermal image and saliency attribute, regardless of human intuition, dataset annotation, or prior assumption. The specific manifestations are: (1) Humans cannot perceive the thermal information of the target through their eyes without the assistance of external equipment; (2) People still mainly based on the RGB image when labeling the RGB-T SOD dataset, unless it is a low-light dark scene; (3) In the RGB-T SOD task, there is no prior hypothesis that the higher the object temperature is, the more salient it is. In other words, thermal image can be misleading and fraudulent. There may be many hot places in the thermal image, but many high-temperature regions are not really salient objects, or it is also possible that the salient object is low temperature while its surrounding environment is high temperature, such as the fourth row of Fig. \ref{fig1}(a). Based on these observations and analyses, one of the first things we do is to enrich the saliency semantic of the thermal features and bridge the gap between thermal modality and saliency task. Therefore, we design a Semantic Constraint Provider (SCP) module in the encoding stage to solve the above problems, as shown in Fig. \ref{fig3}.

As we all know, the top layer of the RGB features contain high-level semantic information, such as the saliency category, which are very important for determining salient objects. Therefore, we can attach the high-level semantic information of the RGB modality to the thermal modality, thereby providing it with more explicit semantic content guidance and strengthening the relevance to the SOD task. First, the top-layer RGB encoder features $E_r^5$ are embedded into a $1\times 1$ convolution layer for channel compression, and then activate it as a semantic mask through a sigmoid function. The concrete process can be described as:
\begin{equation}\label{illuma}
M_r=\delta (Conv_{1\times 1}(E_r^5))
\end{equation}
where $M_r$ is the generated semantic mask, and $Conv_{1\times 1} (\cdot)$ denotes a convolutional layer with the kernel size of $1\times 1$. Note that, in order to ensure the effectiveness of the generated semantic mask, we use the saliency ground truth as the supervision for mask learning. 

With the semantic mask, we can use to guide the thermal modality learning in different situations. Specifically, in normal lighting scenes, that is, when the global illuminance score $\alpha$ is large, the discrimination of salient objects mainly relies on RGB modality, and the generated semantic mask is also more accurate at this time. Thus, we can use the $\alpha$-weighted semantic mask to update the thermal features, thereby suppressing the locations with higher thermal values but not salient, while highlighting the important salient regions of RGB modality in the thermal modality. Similarly, if the brightness of the scene is relatively low, the value of $\alpha$ will be relatively small, and the semantic information extracted from the RGB features may be unreliable. Faced with this situation, our $\alpha$-weighted semantic guidance strategy can still maintain the original thermal features as much as possible and reduce the negative effects of RGB information. We first up-sample (if any) the obtained semantic mask $M_r$ to the size of the thermal encoder features $E_t^i$ through transposed convolution. Then, we multiply the semantic mask $M_r$ by the global illuminance score $\alpha$, thereby generating the $\alpha$-weighted semantic mask. Finally, we use the residual connection to obtain the semantic-embedded thermal encoder features $\hat{E}{_t^i}$. The aforementioned process is defined as:
\begin{equation}\label{illuma}
\hat{E}{_t^i} =\left\{\begin{matrix}
E_t^i+\alpha \cdot (UP(M_r))\otimes E_t^i,i=\left \{ 1,2,3,4 \right \}
 \\
E_t^i+\alpha \cdot M_r\otimes E_t^i,i=5
\end{matrix}\right.
\end{equation}
where $E_t^i$ denote the thermal encoder features of the $i^{th}$ layer, $UP(\cdot)$ is the up-sampling operation, $\alpha$ represents the global illuminance score, and $\otimes$ denotes the element-wise multiplication. 

\begin{figure}[!t]
\centering
\centerline{\includegraphics[width=1\linewidth]{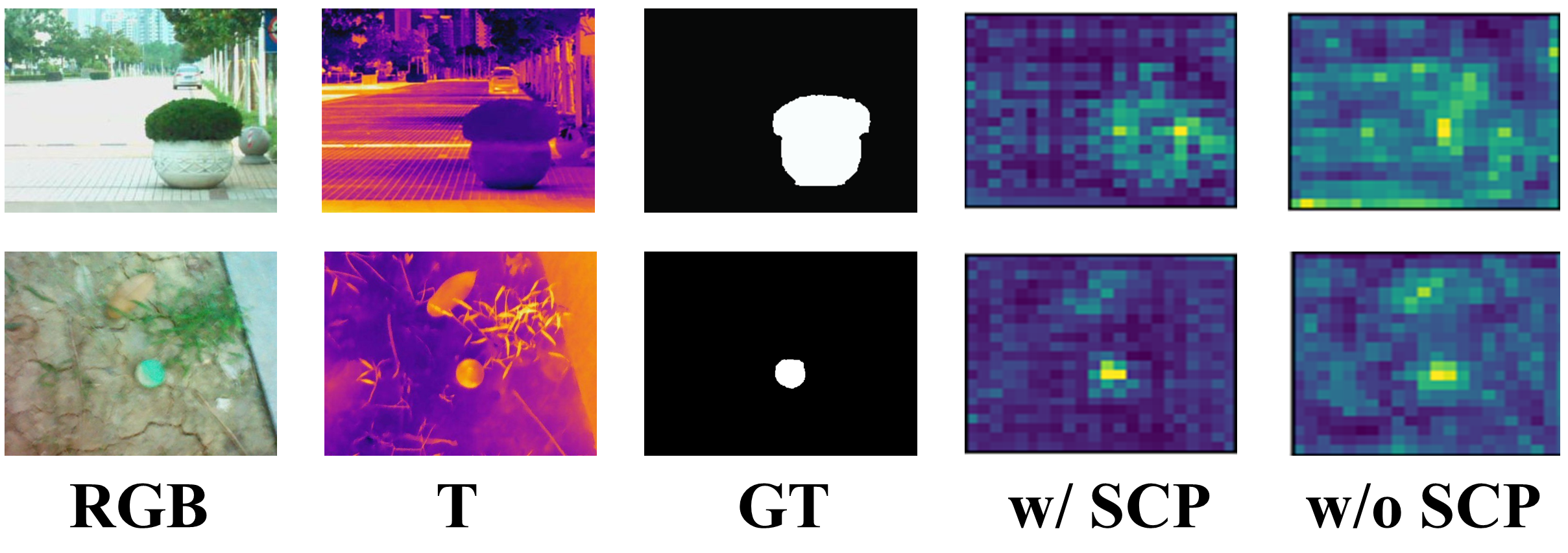}}
\caption{Visual comparison of the features with/without the SCP module.}
\label{fig_rr}
\end{figure}

We present the feature visualization results extracted from the fourth layer of the encoding phase with and without the SCP module in Fig. \ref{fig_rr}, where `w/ SCP' and `w/o SCP' denote model with and without the SCP module, respectively. Compared with the features in fourth and fifth columns, we can see that the SCP module can mask out the misleading information brought by the thermal image. For example, in the first row of the thermal image, the temperature of the salient object is much lower than the ambient temperature, which is obviously very noisy. After adding the SCP module to supplement the semantic information, the salient regions can be correctly highlighted and misleading interference (such as the ground with higher temperature on the left side) can be effectively suppressed. In the second row of the thermal image, the disturbing objects with similar temperatures (such as the leaves above) around the salient object can also be effectively suppressed by the SCP module. Both examples can prove the positive role of the SCP module. More specific ablation experiments can be seen in Section \ref{sec4-ab}.

\subsection{ Localization and Complementation (LC) module}
\begin{figure}[!t]
\centering
\centerline{\includegraphics[width=1\linewidth]{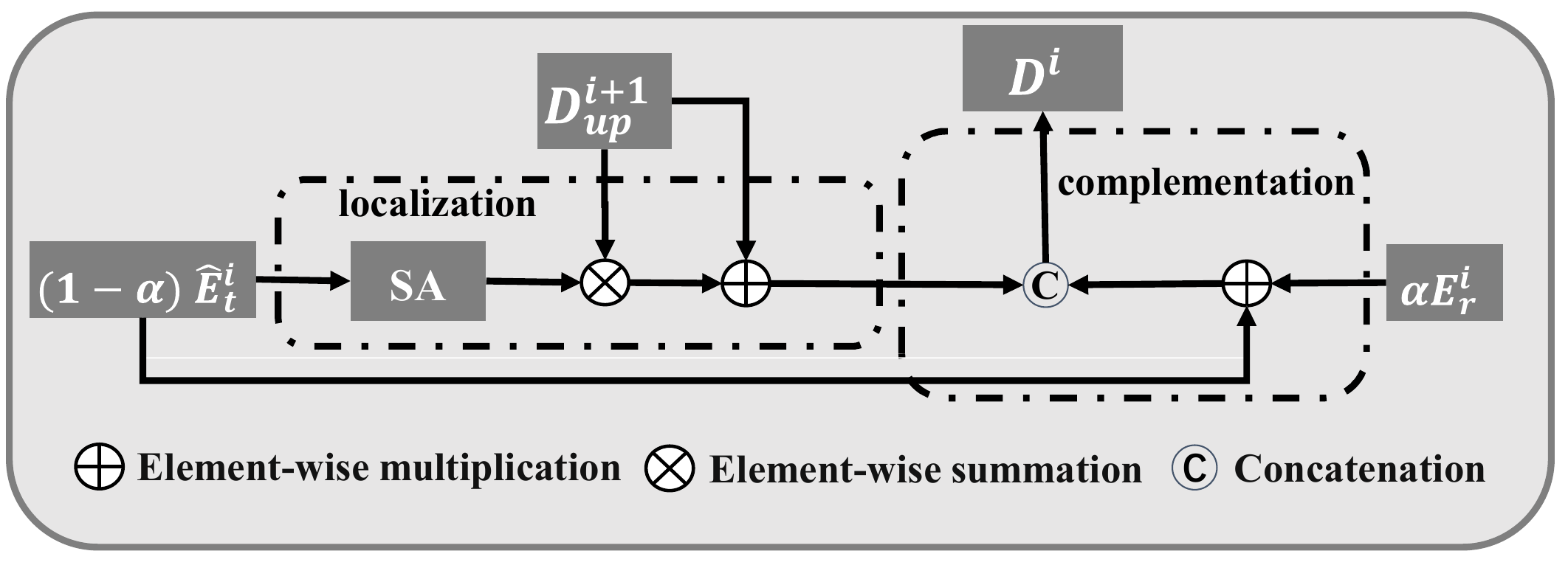}}
\caption{Structure diagram of the Localization and Complementation module.}
\label{fig4}
\end{figure}

Considering the weak correlation between thermal image and saliency attribute, in order to better and more fully play the role of different modality features and obtain a more robust detection result, we mainly rely on RGB features for decoding. There are three main reasons for this. First, according to the setting of salient objects in RGB-T images, people mainly judge whether objects are salient or not based on RGB image, while thermal image is more to provide some auxiliary information to deal with some challenging scenes. For example, people are still mainly based on RGB image when annotating the RGB-T SOD dataset, except for some low-light dark scenes. Second, starting from the human intuition and definition of salient object, humans cannot perceive the heat information of the target through their eyes without the assistance of external devices, but salient objects are the most attractive objects in a given scene, so there is a weak correlation between thermal image and saliency attribute. Third, there is no a priori assumption that objects with higher temperatures are more salient. For these reasons, we prefer thermal image to play an auxiliary role in the decoding process to avoid problems that may be caused by overemphasizing thermal features (such as incomplete or wrong detection). Therefore, we set up a single-stream decoding branch dominated by RGB features and assisted by thermal features. 
Concretely, we design the Localization and Complementation (LC) module to complete progressive feature decoding. As shown in Fig. \ref{fig4}, the LC module is decoupled into two stages, that is, the localization guidance stage and the skip-connection complementation stage.

\begin{figure*}[h]
	\begin{minipage}{0.333\linewidth}
		\vspace{3pt}
		\centerline{\includegraphics[width=\textwidth]{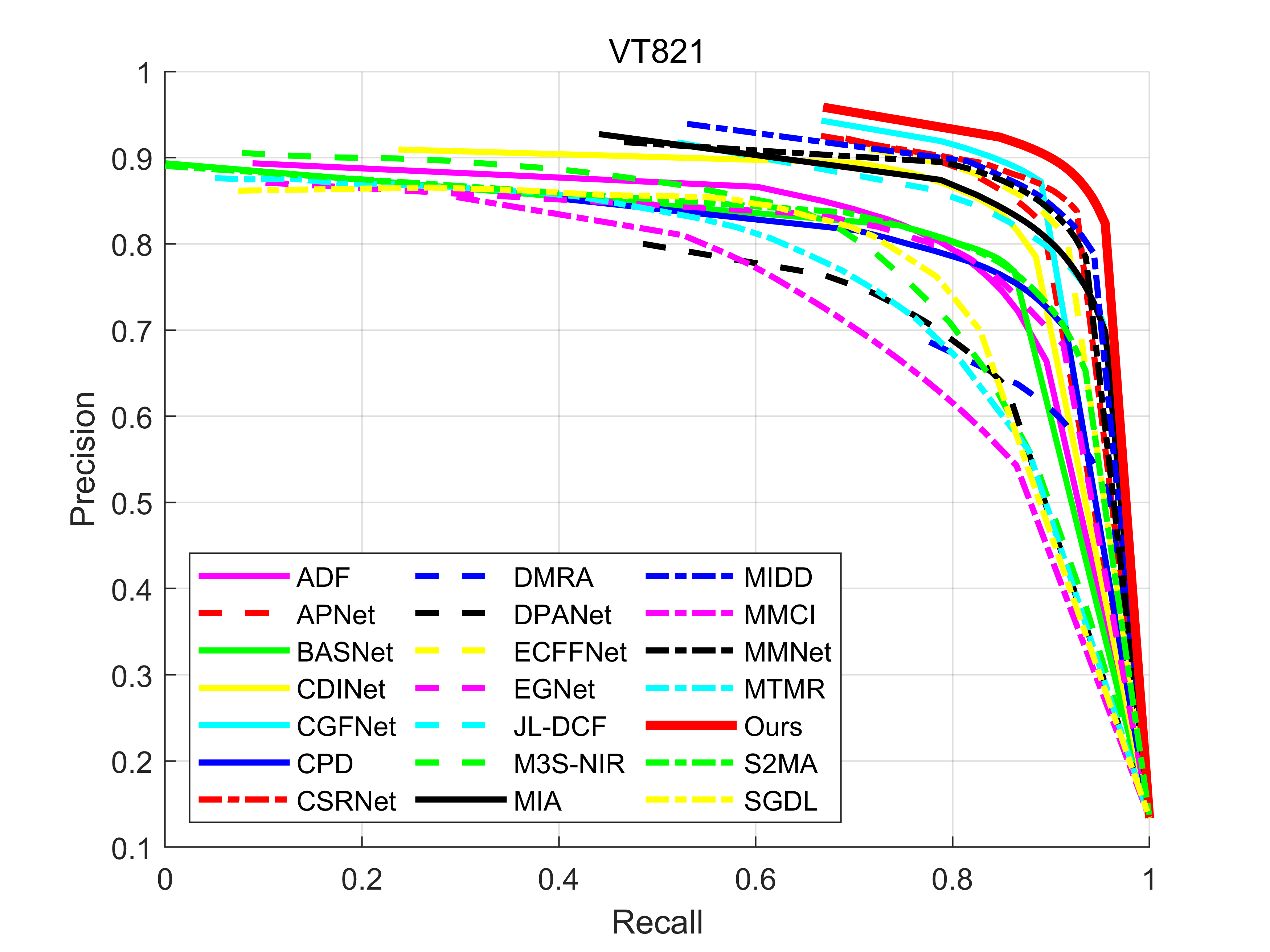}}
	\end{minipage}
	\begin{minipage}{0.333\linewidth}
		\vspace{3pt}
		\centerline{\includegraphics[width=\textwidth]{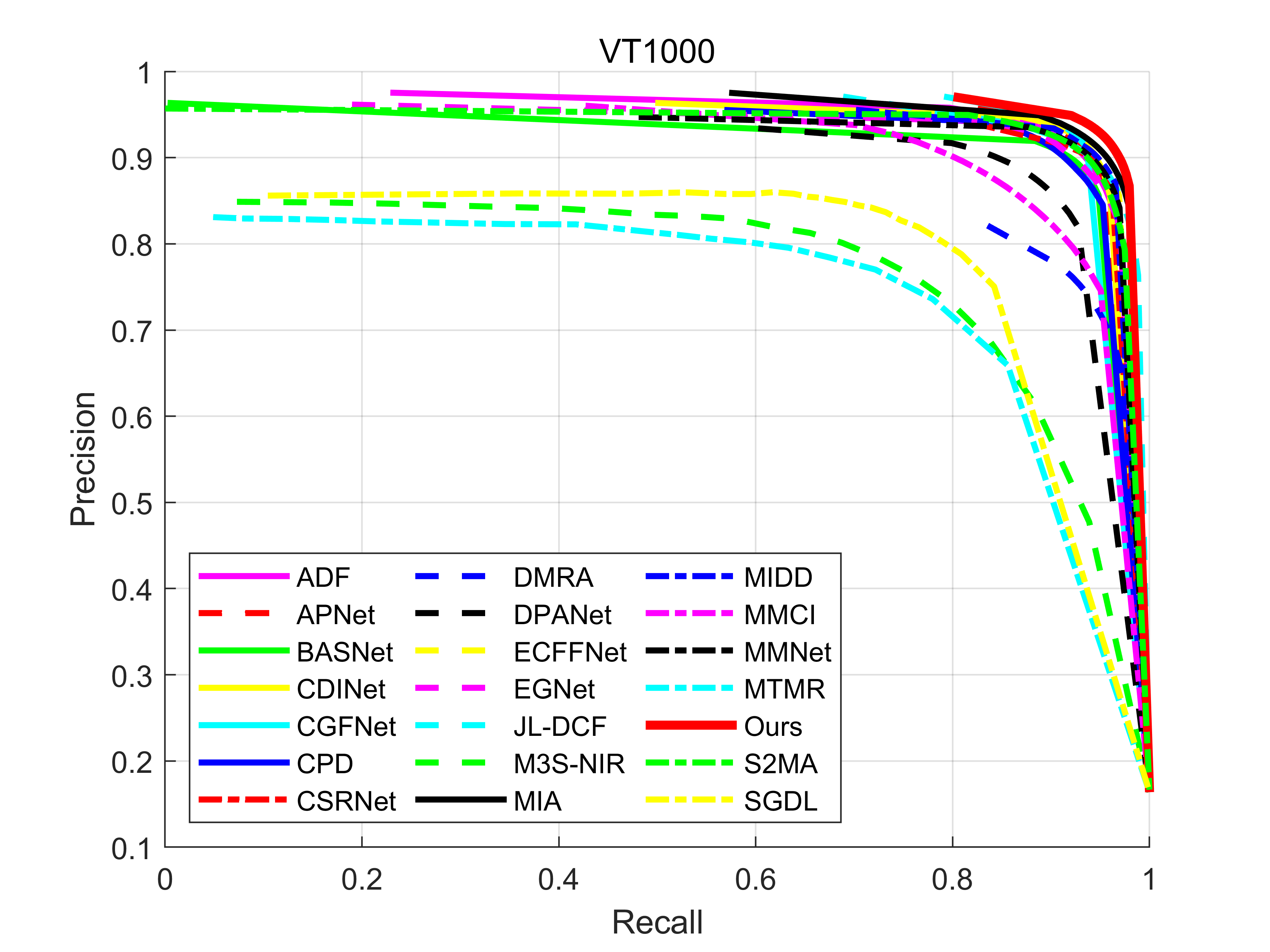}}
	\end{minipage}
	\begin{minipage}{0.333\linewidth}
		\vspace{3pt}
		\centerline{\includegraphics[width=\textwidth]{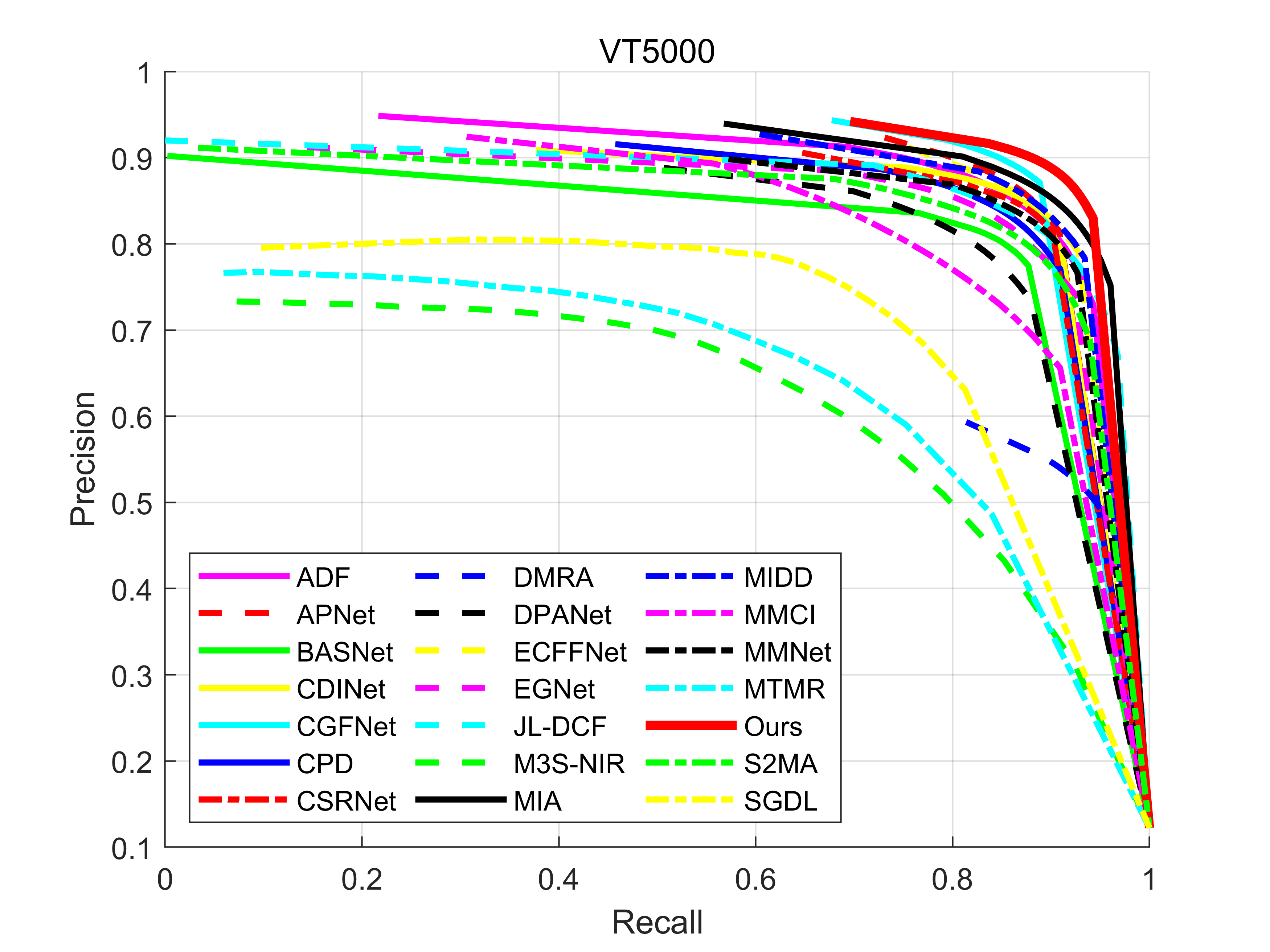}}
	\end{minipage}
	\caption{P-R curves of different methods on the VT821, VT1000, and VT5000 datasets. }
	\label{pr}
\end{figure*}

First, accurate localization of salient objects is an important part in SOD task, but it is also a very challenging problem. Generally speaking, localization of salient objects can be done by RGB image, but in some difficult scenes, relying on RGB image alone is insufficient. In the RGB-T SOD task, thermal image is introduced as an auxiliary modality to help detect the salient objects, so we need to consider how to mine thermal information to assist object localization. According to our previous analysis of thermal image properties, we semantically complement thermal features with the help of the SCP module during the encoding phase, bridging the correlation gap between thermal image and saliency attribute. Although the enhanced thermal feature may still not be perfect in terms of regional integrity, it can still provide some valuable location information, which is beneficial to our RGB-based feature decoding. Based on this, we design a localization guidance stage in the LC model by learning a weight map to provide localization information from thermal modality. 
Moreover, considering that the localization effect of thermal features should play a greater role in low-light scenes, the global illuminance score $\alpha$ is also introduced in the localization guidance stage to control the role of thermal features. We first multiply $1-\alpha$ with the semantic-embedded thermal encoder features $\hat{E}{_t^i}$, introducing the effects of different lighting factors. Then, we use spatial attention to adaptively obtain a weight map reflecting the importance of spatial locations in the thermal features. After that, we use it to modify the decoder features of the previous layer. The location-refined decoder features $D_L^i$ can be expressed as:
\begin{equation}\label{illuma}
D_L^i=D^{i+1}_{up}+SA((1-\alpha )\cdot\hat{E}{_t^i})\otimes D^{i+1}_{up}
\end{equation}
where $SA(\cdot)$ represents the spatial attention operation \cite{woo2018cbam,crm/tim22/covid}, $D^{i+1}_{up}$ denote $2\times$ up-sampled decoder features from the $i+1$ level by using the `Up' block that consists of one $2\times$ linear interpolation and two convolutional layers, where each convolutional layer is followed by a batch normalization and a ReLU activation.

In addition, skip connection features have been shown to be effective in the SOD task \cite{pang2020hierarchical,crm/tcsvt22/weaklySOD,crm/nips20/CoADNet,crm/tcyb22/glnet,crm/tgrs19/rsi}. Considering this, in the skip-connection complementation stage, we adaptively fuse the RGB and thermal encoder features through the global illuminance score to generate more sufficient and comprehensive skip connection information, which is further used to complement the decoder features. The skip connection features can be expressed as:
\begin{equation}\label{illuma}
F_c^i=\alpha \cdot E_r^i+(1-\alpha )\cdot \hat{E}{_t^i}
\end{equation}
Finally, the location-refined decoder features $D_L^i$ and skip connection features $F_c^i$ are concatenated, and then use a simple channel attention mechanism \cite{hu2018squeeze} to emphasize more useful channels and get the the final decoder feature of this layer:
\begin{equation}\label{illuma}
D^i=Conv_{1\times 1}(CA(Cat(F_c^i,D_L^i)))
\end{equation}
where $CA(\cdot)$ represents the channel attention operation \cite{hu2018squeeze,crm/tim22/covid}, $Cat(\cdot,\cdot)$ denotes channel-wise concatenation operation, and $Conv_{1\times 1}$ is used to reduce the number of channels through a convolutional layer with the kernel size of $1\times 1$.


We reduce the number of feature map channels obtained by decoding the last layer to one channel through a convolutional layer with the kernel size of $1\times 1$, adjust the resolution of the feature map to the size of the input image by interpolation and up-sampling, and then obtain the final saliency map through sigmoid activation.

\subsection{Loss Function}
During training, we use the BCE loss \cite{crm/tip22/CIRNet} and IoU loss \cite{mattyus2017deeproadmapper} as the supervision for the saliency map $D^i$ and semantic mask $M_r$. 
The final loss function is defined as:
\begin{equation}\label{illuma}
\begin{aligned}
\ell _{total}&=\sum_{i=1}^{5}( \ell _{bce}(D^i,G)+\ell _{IoU}(D^i,G))\\ &+  \ell _{bce}(M_r,G)+\ell _{IoU}(M_r,G)    
\end{aligned}
\end{equation}
where $G$ denotes the saliency ground truth, $\ell _{bce}$ and $\ell _{IoU}$ are the BCE loss and IoU loss, respectively.


\section{Experiments} \label{sec4}

\subsection{Datasets and Evaluation Metrics}

We use three publicly available RGT-T SOD benchmark datasets to demonstrate the effectiveness of the proposed method, including VT821 dataset \cite{wang2018rgb}, VT1000 dataset \cite{tu2019rgb}, and VT5000 dataset \cite{9767629}. The VT821 dataset contains 821 image pairs aligned in different scenes, since the RGB-T images of the VT821 dataset are manually registered, so missing areas will appear in thermal images. The VT1000 dataset includes 1,000 RGB-T image pairs and the corresponding pixel-wise saliency ground truth. The VT5000 dataset is the largest RGB-T SOD dataset, containing 5,000 pairs of RGB-T images, including multiple complex scenes. In the experiments, we use the 2,500 pairs in the VT5000 dataset as the training set, the remaining 2,500 pairs in the VT5000, VT1000, and VT821 datasets as the testing set. 
We employ five widely used metrics to evaluate the performance of each model, including precision-recall (P-R) curves\cite{crm/tip18/RGBDCoSOD,crm/tmm19/HSCS}, maximum F-measure ($F_{\beta}^m$) \cite{achanta2009frequency,crm/tip19/VSOD,crm/ICME18/CoSOD}, Mean Absolute Error ($MAE$) \cite{crm/tcyb20/going,crm/tcyb19/iterativeCoSOD}, S-measure ($S_m$)\cite{fan2017structure}, and E-measure ($E_\varphi$) \cite{fan2018enhanced}.

\begin{table*}[t]
\caption{Quantitative comparison results of different models. $\uparrow$/$\downarrow$ for a metric denotes that a larger/smaller value is better.The best results are in bold and the second best results are highlighted in underline}
\resizebox{\linewidth}{!}{
\begin{tabular}{c|c|c|cccc|cccc|cccc}
\hline
\multirow{2}{*}{Type}                     & \multirow{2}{*}{Model} &\multirow{2}{*}{From} & \multicolumn{4}{c|}{VT5000}                                                                                                      & \multicolumn{4}{c|}{VT1000}                                                                                                     & \multicolumn{4}{c}{VT821}                                                                                                       \\ \cline{4-15} 
                                         &                         & & \multicolumn{1}{c}{$S_m\uparrow $}              & \multicolumn{1}{c}{$E_\varphi\uparrow $}              & \multicolumn{1}{c}{$F_{\beta}^m\uparrow $}           & $MAE\downarrow$            & \multicolumn{1}{c}{$S_m\uparrow $}              & \multicolumn{1}{c}{$E_\varphi\uparrow $}              & \multicolumn{1}{c}{$F_{\beta}^m\uparrow $}           & $MAE\downarrow$           & \multicolumn{1}{c}{$S_m\uparrow $}              & \multicolumn{1}{c}{$E_\varphi\uparrow $}              & \multicolumn{1}{c}{$F_{\beta}^m\uparrow $}           & $MAE\downarrow$            \\ \hline
\multirow{3}{*}{RGB SOD}               & CPD &ICCV'19                    & \multicolumn{1}{c}{0.855}          & \multicolumn{1}{c}{0.894}          & \multicolumn{1}{c}{0.847}          & 0.046          & \multicolumn{1}{c}{0.907}          & \multicolumn{1}{c}{0.923}          & \multicolumn{1}{c}{0.914}          & 0.031         & \multicolumn{1}{c}{0.818}          & \multicolumn{1}{c}{0.843}          & \multicolumn{1}{c}{0.786}          & 0.079          \\ 
                                          & EGNet    &ICCV'19               & \multicolumn{1}{c}{0.853}          & \multicolumn{1}{c}{0.888}          & \multicolumn{1}{c}{0.840}           & 0.051          & \multicolumn{1}{c}{0.910}           & \multicolumn{1}{c}{0.922}          & \multicolumn{1}{c}{0.917}          & 0.033         & \multicolumn{1}{c}{0.830}           & \multicolumn{1}{c}{0.856}          & \multicolumn{1}{c}{0.795}          & 0.063          \\ 
                                          & BASNet     &CVPR'19             & \multicolumn{1}{c}{0.839}          & \multicolumn{1}{c}{0.878}          & \multicolumn{1}{c}{0.821}          & 0.054          & \multicolumn{1}{c}{0.909}          & \multicolumn{1}{c}{0.923}          & \multicolumn{1}{c}{0.913}          & 0.030          & \multicolumn{1}{c}{0.823}          & \multicolumn{1}{c}{0.856}          & \multicolumn{1}{c}{0.803}          & 0.067          \\ \hline
\multirow{7}{*}{RGB-D SOD}             & MMCI      &PR'19              & \multicolumn{1}{c}{0.819}          & \multicolumn{1}{c}{0.857}          & \multicolumn{1}{c}{0.797}          & 0.056          & \multicolumn{1}{c}{0.879}          & \multicolumn{1}{c}{0.891}          & \multicolumn{1}{c}{0.876}          & 0.040          & \multicolumn{1}{c}{0.759}          & \multicolumn{1}{c}{0.784}          & \multicolumn{1}{c}{0.723}          & 0.089          \\ 
                                          
                                          & DMRA     &ICCV'19               & \multicolumn{1}{c}{0.659}          & \multicolumn{1}{c}{0.666}          & \multicolumn{1}{c}{0.632}          & 0.184          & \multicolumn{1}{c}{0.784}          & \multicolumn{1}{c}{0.801}          & \multicolumn{1}{c}{0.824}          & 0.124         & \multicolumn{1}{c}{0.666}          & \multicolumn{1}{c}{0.691}          & \multicolumn{1}{c}{0.702}          & 0.216          \\ 
                                          & S2MA       &CVPR'20             & \multicolumn{1}{c}{0.854}          & \multicolumn{1}{c}{0.864}          & \multicolumn{1}{c}{0.833}          & 0.054          & \multicolumn{1}{c}{0.919}          & \multicolumn{1}{c}{0.912}          & \multicolumn{1}{c}{0.922}          & 0.030          & \multicolumn{1}{c}{0.812}          & \multicolumn{1}{c}{0.813}          & \multicolumn{1}{c}{0.804}          & 0.098          \\ 
                                          & JL-DCF     &CVPR'20             & \multicolumn{1}{c}{0.862}          & \multicolumn{1}{c}{0.860}           & \multicolumn{1}{c}{0.850}           & 0.050           & \multicolumn{1}{c}{0.913}          & \multicolumn{1}{c}{0.899}          & \multicolumn{1}{c}{0.922}          & 0.030          & \multicolumn{1}{c}{0.839}          & \multicolumn{1}{c}{0.830}           & \multicolumn{1}{c}{0.844}          & 0.076          \\ 
                                          & DPANet     &TIP'21             & \multicolumn{1}{c}{0.860}           & \multicolumn{1}{c}{0.865}          & \multicolumn{1}{c}{0.832}          & 0.050           & \multicolumn{1}{c}{0.881}          & \multicolumn{1}{c}{0.881}          & \multicolumn{1}{c}{0.863}          & 0.051         & \multicolumn{1}{c}{0.736}          & \multicolumn{1}{c}{0.735}          & \multicolumn{1}{c}{0.715}          & 0.157          \\ 
                                          & CDINet    &ACM MM'21              & \multicolumn{1}{c}{0.867}          & \multicolumn{1}{c}{0.903}          & \multicolumn{1}{c}{0.860}          & 0.044          & \multicolumn{1}{c}{0.920}           & \multicolumn{1}{c}{0.931}          & \multicolumn{1}{c}{0.930}           & 0.026         & \multicolumn{1}{c}{0.844}          & \multicolumn{1}{c}{0.880}           & \multicolumn{1}{c}{0.849}          & 0.056          \\ \hline
                                          
\multirow{3}{*}{\begin{tabular}[c]{@{}c@{}}Traditional \\ RGB-T SOD\end{tabular}} & SGDL  &TMM'20                  & \multicolumn{1}{c}{0.750}           & \multicolumn{1}{c}{0.824}          & \multicolumn{1}{c}{0.737}          & 0.089          & \multicolumn{1}{c}{0.787}          & \multicolumn{1}{c}{0.856}          & \multicolumn{1}{c}{0.807}          & 0.090          & \multicolumn{1}{c}{0.764}          & \multicolumn{1}{c}{0.846}          & \multicolumn{1}{c}{0.780}           & 0.085          \\ 
                                          & M3S    &MIPR'20                 & \multicolumn{1}{c}{0.652}          & \multicolumn{1}{c}{0.780}           & \multicolumn{1}{c}{0.644}          & 0.168          & \multicolumn{1}{c}{0.726}          & \multicolumn{1}{c}{0.827}          & \multicolumn{1}{c}{0.769}          & 0.145         & \multicolumn{1}{c}{0.723}          & \multicolumn{1}{c}{0.859}          & \multicolumn{1}{c}{0.780}           & 0.140           \\ 
                                          & MTMR    &IGTA'18                & \multicolumn{1}{c}{0.680}           & \multicolumn{1}{c}{0.795}          & \multicolumn{1}{c}{0.662}          & 0.114          & \multicolumn{1}{c}{0.706}          & \multicolumn{1}{c}{0.836}          & \multicolumn{1}{c}{0.755}          & 0.119         & \multicolumn{1}{c}{0.725}          & \multicolumn{1}{c}{0.815}          & \multicolumn{1}{c}{0.747}          & 0.108          \\ \hline
                                          
\multirow{9}{*}{\begin{tabular}[c]{@{}c@{}}CNNs-based \\ RGB-T SOD\end{tabular}}  
& APNet      &TCSVT'21             & \multicolumn{1}{c}{0.875}          & \multicolumn{1}{c}{0.914}          & \multicolumn{1}{c}{0.875}          & {\ul 0.035}    & \multicolumn{1}{c}{0.921}          & \multicolumn{1}{c}{\ul 0.937}          & \multicolumn{1}{c}{0.929}          & \textbf{0.021}   & \multicolumn{1}{c}{0.867}          & \multicolumn{1}{c}{0.907}          & \multicolumn{1}{c}{0.865}          & 0.034          \\ 
& MIDD       &TIP'21             & \multicolumn{1}{c}{0.874}          & \multicolumn{1}{c}{0.897}          & \multicolumn{1}{c}{0.871}          & 0.038          & \multicolumn{1}{c}{0.915}          & \multicolumn{1}{c}{0.933}          & \multicolumn{1}{c}{0.926}          & 0.027         & \multicolumn{1}{c}{0.871}          & \multicolumn{1}{c}{0.895}          & \multicolumn{1}{c}{0.874}          & 0.045          \\ 
& MIA       &NC'22              & \multicolumn{1}{c}{0.879}          & \multicolumn{1}{c}{0.893}          & \multicolumn{1}{c}{0.880}           & 0.040           & \multicolumn{1}{c}{{\ul 0.924}}    & \multicolumn{1}{c}{0.926}          & \multicolumn{1}{c}{{\ul 0.935}}    & 0.025         & \multicolumn{1}{c}{0.844}          & \multicolumn{1}{c}{0.850}           & \multicolumn{1}{c}{0.854}          & 0.070           \\ 
& ADF    &TMM'22                 & \multicolumn{1}{c}{0.864}          & \multicolumn{1}{c}{0.891}          & \multicolumn{1}{c}{0.863}          & 0.048          & \multicolumn{1}{c}{0.910}           & \multicolumn{1}{c}{0.921}          & \multicolumn{1}{c}{0.923}          & 0.034         & \multicolumn{1}{c}{0.810}           & \multicolumn{1}{c}{0.843}          & \multicolumn{1}{c}{0.804}          & 0.077          \\ 
                                          & ECFFNet   &TCSVT'22              & \multicolumn{1}{c}{0.874}          & \multicolumn{1}{c}{0.906}          & \multicolumn{1}{c}{0.872}          & 0.038          & \multicolumn{1}{c}{0.923}          & \multicolumn{1}{c}{0.930}           & \multicolumn{1}{c}{0.930}           & \textbf{0.021}   & \multicolumn{1}{c}{0.877}          & \multicolumn{1}{c}{0.902}          & \multicolumn{1}{c}{0.865}          & 0.034          \\ 
                                          & MMNet      &TCSVT'22             & \multicolumn{1}{c}{0.864}          & \multicolumn{1}{c}{0.890}           & \multicolumn{1}{c}{0.852}          & 0.042          & \multicolumn{1}{c}{0.917}          & \multicolumn{1}{c}{0.924}          & \multicolumn{1}{c}{0.920}           & 0.027         & \multicolumn{1}{c}{0.875}          & \multicolumn{1}{c}{0.893}          & \multicolumn{1}{c}{0.867}          & 0.040           \\ 
                                          & CSRNet     &TCSVT'22             & \multicolumn{1}{c}{0.868}          & \multicolumn{1}{c}{0.905}          & \multicolumn{1}{c}{0.857}          & 0.042          & \multicolumn{1}{c}{0.918}          & \multicolumn{1}{c}{0.925}          & \multicolumn{1}{c}{0.918}          & 0.024         & \multicolumn{1}{c}{{\ul 0.885}}    & \multicolumn{1}{c}{{\ul 0.909}}    & \multicolumn{1}{c}{0.880}           & 0.038          \\ 
                                          & CGFNet     &TCSVT'22             & \multicolumn{1}{c}{{\ul 0.882}}    & \multicolumn{1}{c}{{\ul 0.921}}    & \multicolumn{1}{c}{{\ul 0.888}}    & {\ul 0.035}    & \multicolumn{1}{c}{0.921}          & \multicolumn{1}{c}{\textbf{0.941}} & \multicolumn{1}{c}{0.933}          & {{\ul 0.023}}         & \multicolumn{1}{c}{0.879}          & \multicolumn{1}{c}{0.912}          & \multicolumn{1}{c}{{\ul 0.883}}    & {\ul 0.036}    \\ 

                                          & Ours      &-               & \multicolumn{1}{c}{\textbf{0.895}} & \multicolumn{1}{c}{\textbf{0.927}} & \multicolumn{1}{c}{\textbf{0.895}} & \textbf{0.033} & \multicolumn{1}{c}{\textbf{0.929}} & \multicolumn{1}{c}{{\ul 0.937}}    & \multicolumn{1}{c}{\textbf{0.937}} & \textbf{0.021} & \multicolumn{1}{c}{\textbf{0.899}} & \multicolumn{1}{c}{\textbf{0.919}} & \multicolumn{1}{c}{\textbf{0.904}} & \textbf{0.030} \\ \hline
\end{tabular}}
\label{tab1}
\end{table*}

\begin{figure*}[!t]
\centering
\centerline{\includegraphics[width=1\linewidth]{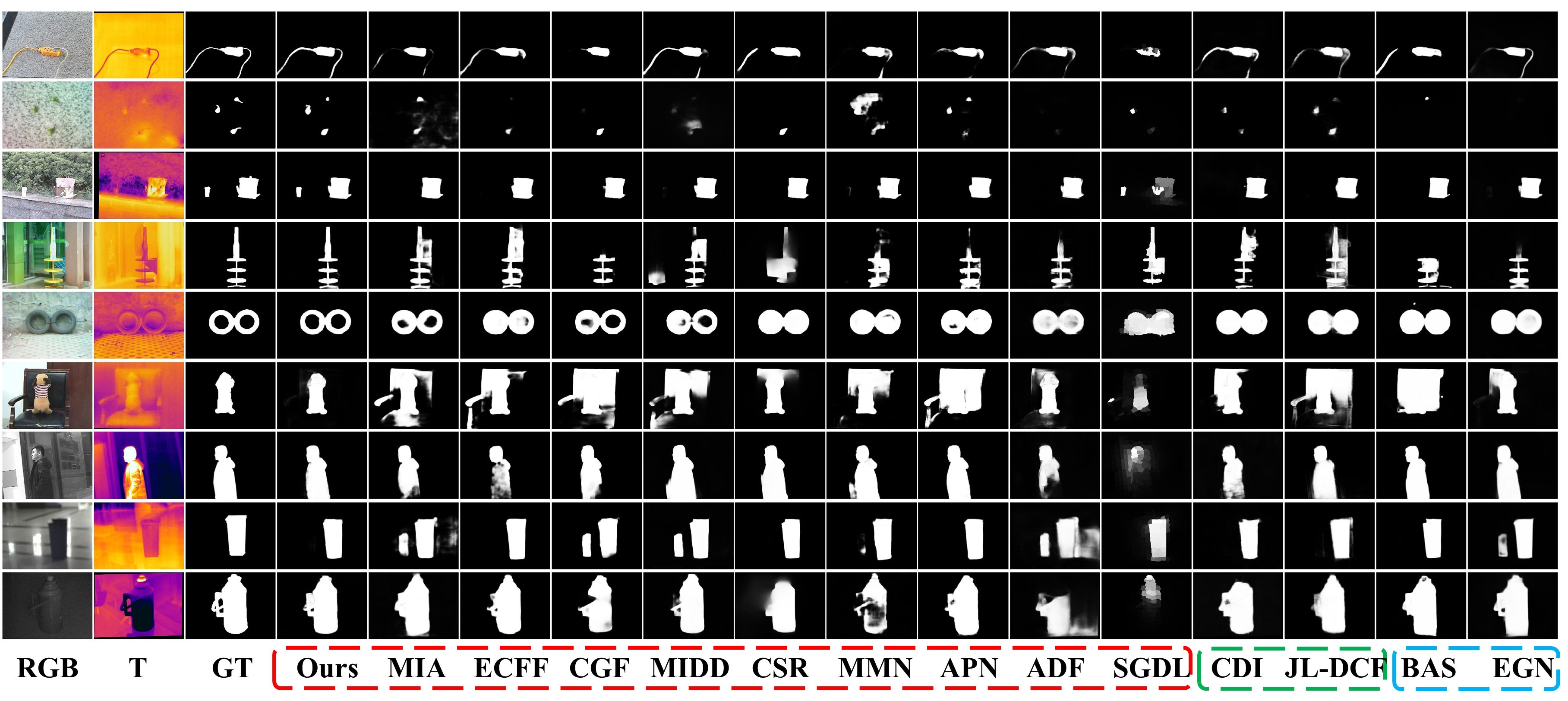}}
\caption{Qualitative visual comparison of different methods, including nine state-of-the-art RGB-T SOD methods marked in red dotted box, two RGB-D SOD methods marked in green dotted box, and two RGB SOD methods marked in blue dotted box.}
\label{fig5}
\end{figure*}
\subsection{Implementation details}
We implement our model using the PyTorch toolbox with an NVIDIA GeForce RTX 3090 GPU. We also implement our network by using the MindSpore Lite tool\footnote{\url{https://www.mindspore.cn/}}. The backbone network is initialized with parameters pre-trained on ImageNet\cite{russakovsky2015imagenet}, and other parameters are initialized with default PyTorch settings. The training samples are augmented with multiple strategies, including random flipping, rotation, boundary clipping, adding probabilistic noise, and multi-scale input. We use the Adam \cite{kingma2014adam} optimizer to train our model with an initial learning rate set to 1e-4 and dividing the learning rate by 10 every 45 epochs. We train 100 epochs with the batch size of 16. All input images are resized to $352\times 352$ for testing.
During training, only BCE loss is used for the first 30 epochs, and then IoU loss is added to further supervise globally.

\subsection{Comparison with State-of-the-Art Methods}
To verify the superiority of the proposed model, we compare it with twenty state-of-the-art (SOTA) SOD methods, which can be further classified into four categories: (1) three RGB SOD methods including CPD\cite{wu2019cascaded}, BASNet\cite{qin2019basnet} , and EGNet\cite{zhao2019egnet}; (2) six RGB-D SOD methods including DMRA\cite{piao2019depth}, MMCI\cite{CHEN2019376}, JL-DCF\cite{fu2020jl}, S2MA\cite{liu2020learning}, DPANet\cite{crm/tip21/DPANet}, and CDINet\cite{crm/acmmm21/CDINet}; (3) three traditional RGB-T SOD methods including SGDL\cite{tu2019rgb}, MTMR\cite{wang2018rgb}, and M3S-NIR\cite{tu2019m3s}; (4) eight deep learning RGB-T SOD methods including ADF\cite{9767629}, MMNet\cite{9439490}, MIDD\cite{tu2021multi}, APNet\cite{9583676}, ECFFNet\cite{9420662}, CSRNet\cite{9505635}, CGFNet\cite{9493207}, and MIA\cite{liang2022multi}.
All RGB and RGB-D SOD models are retrained on the same RGB-T training dataset as our model for fair comparison.

\subsubsection{Qualitative Comparison}
For quantitative evaluation, P-R curves of all compared methods on three benchmark datasets are shown in Fig. \ref{pr}. The closer the P-R curve is to (1,1), the better the performance. From it, we can see that the proposed method achieves higher precision and recall scores on all datasets compared to other competitors. 
A more intuitive numerical comparison is presented in Table \ref{tab1}.

First, it is clear that compared with other deep learning-based RGB-T SOD methods, the results of deep learning-based RGB SOD methods are not satisfactory because the thermal modality is not exploited. Second, the closest class of methods to RGB-T SOD algorithms are RGB-D SOD models, but it is observed that they generally cannot exceed the performance of methods dedicated to RGB-T SOD task. For example, the state-of-the-art RGB-D SOD methods (such as CDINet \cite{crm/acmmm21/CDINet}) have been shown to work well on the RGB-D SOD datasets, but even if they were retrained on the VT5000 training set, they still perform poorly on the RGB-T SOD dataset. This is mainly due to the difference between depth map and thermal image. We all know that the depth map has a strong correlation with saliency attribute, and the depth of scene information can be directly perceived by people, which in turn directly affects people's recognition of salient objects. Therefore, in the model design of existing RGB-D SOD methods, there are usually explicit depth information guidance or priors that can be utilized (such as, the larger the depth value, the more salient the object, \etc). However, the thermal image is weakly correlated with saliency because it is difficult for humans to perceive thermal radiation from objects without external equipment, and there is no prior assumption that objects with higher temperature are more salient. Therefore, directly applying the RGB-D SOD method to the RGB-T SOD task may results in poor performance. Moreover, even some RGB-D SOD methods perform worse than RGB SOD methods. For example, the MMCI method \cite{CHEN2019376} does not outperform the contemporaneous CPD method \cite{wu2019cascaded} on these three datasets.

For the RGB-T SOD methods, the performance of traditional methods is far inferior to deep learning-based methods including the RGB and RGB-D methods, mainly due to the limited feature representation ability. Among the deep learning-based RGB-T SOD methods, our proposed TNet can achieve the best performance on all metrics, except the E-measure on the VT1000 dataset. For example, compared with the second best method on the VT821 dataset, our method wins the percentage gain of 16.7\% for MAE score, 1.6\% for S-measure, 2.4\% for F-measure, and 1.1\% for E-measure, respectively. On the VT5000 dataset, the minimum percentage gain reaches 5.7\% in terms of MAE score, 1.5\% in terms of S-measure, and 0.8\% in terms of F-measure, respectively. On the VT1000 dataset, compared with the latest CNNs-based RGB-T SOD methods, the maximum percentage gain of MAE score reaches 38.2\%. The E-measure of our proposed TNet ranks second on the VT1000 dataset, and the performance improvements of other metrics are not as significant as on other datasets. This is because the scenes in the VT1000 dataset are relatively simple, and most of them are in the daytime, so it does not reflect the advantages of our TNet in detecting scenes with different brightness. In fact, for the RGB-T SOD task, the original intention of introducing the thermal modality is to solve the SOD task in low-light environment, so building a large-scale RGB-T SOD dataset containing more low-light scenes is also the focus of our future research. Overall, our method achieves competitive performance, which demonstrates its effectiveness.

\begin{table}[!t]
\centering
\small
\renewcommand\arraystretch{1.4}
    \caption{Comparison of complexity and performance on the VT1000 dataset.}
    \label{table_com}
    \setlength{\tabcolsep}{2.9mm}{
        \begin{tabular}{c|c|c|c}
    \hline
        Model & From & FLOPs & $F_{\beta}^m $ \\ \hline
        CGFNet & TCSVT'22 & 139.97G & 0.933  \\ \hline
        APNet & TCSVT'21 & 46.48G & 0.929  \\ \hline
        MIDD & TIP'21 & 57.44G & 0.926  \\ \hline
        ADF & TMM'22 & 474.71G & 0.923  \\ \hline
        TNet (Ours) & --- & 39.71G & 0.937  \\ \hline
        \end{tabular}}
\end{table}

\subsubsection{Quantitative Comparison}
Some visual comparisons are shown in Fig. \ref{fig5}, including some challenging scenarios, such as small objects, multiple objects, complex backgrounds, low light, and hollow objects.
It can be seen that our method excels in localization accuracy, background suppression, detail representation, and robustness to some challenging scenarios.
For example, as shown in the first three rows of Fig. \ref{fig5}, the scene contains multiple salient objects, slender objects or small objects, and the low color contrast of the scene, which undoubtedly increases the difficulty of detection. For these images, almost all methods are at a loss, while our proposed method locates salient objects accurately and suppresses the interference of irrelevant noise effectively.
Moreover, the detail characterization ability of our method is significantly better than other methods. For example, the hollow objects in the fifth image of Fig. \ref{fig5} and the kettle handle in the last image are all effectively detected.
The last three images in Fig. \ref{fig5} are the low-light scenes, it can be seen that our method achieves better prediction than other SOTA methods. This is because we regulate the interaction between the RGB image and the thermal image through the learned global illuminance score, supplement the semantic content for each layer of the thermal image branch in the encoding stage, and take full advantage of the valuable information that the thermal modality can provide in the decoding stage.


subsubsection{Computational Complexity}
To compare the complexity of different algorithms, we select four open-source RGB-T SOD models for comparison, including CGFNet \cite{9493207}, APNet \cite{9583676}, ADF \cite{9767629}, MIDD \cite{tu2021multi}. Table \ref{table_com} shows the FLOPs (Floating Point Operations) and maximum F-measure of different algorithms on the VT1000 dataset. The smaller the value of FLOPs, the lower the computational complexity. Compared with other methods, our proposed TNet achieves superior performance with only 39.71G FLOPs compared to the second best method in performance (\ie, CGFNet \cite{9493207}). Therefore, our proposed TNet achieves a win-win in detection performance and model complexity.

\subsection{Ablation Study}\label{sec4-ab}

To verify the effectiveness of key components in our model, we conduct experiments by removing or replacing them from our full implementation. The quantitative results are shown in Table \ref{tab2}. What's more, to better understand the role of the key components, we visualize the results of the ablation experiments in Fig. \ref{fig7}.

\subsubsection{Effectiveness of thermal modality}
 To demonstrate the prediction effect of the network with only RGB image, we add an additional ablation experiment denoted `Only RGB'. In fact, if the network only relies on the RGB image, it degenerates into a simple U-Net network, that is, the thermal branch on the right side of the network diagram and the SCP module are all removed, and the LC module is replaced with skip connections with only the RGB branch and no global illuminance score control. The quantitative comparisons are shown in the second row of Table \ref{tab2}. It can be seen that the performance of using only RGB branch is obviously reduced compared to the full model. For example, on the VT821 dataset, compared with model using only RGB image, the F-measure score of the full model is improved from 0.851 to 0.904 with a percentage gain of 6.2\%, and the MAE score also achieves a percentage gain of 41.2\%. It can also be seen from the visualization results shown in Fig. \ref{fig7} that many background noises are not well suppressed using only RGB image (such as the stone on the right in the first image and the white wall in the second image). All these experiments verify the effectiveness of the thermal modality.

\subsubsection{Effectiveness of GIE module}
\begin{figure*}[!t]
\centering
\centerline{\includegraphics[width=1\linewidth]{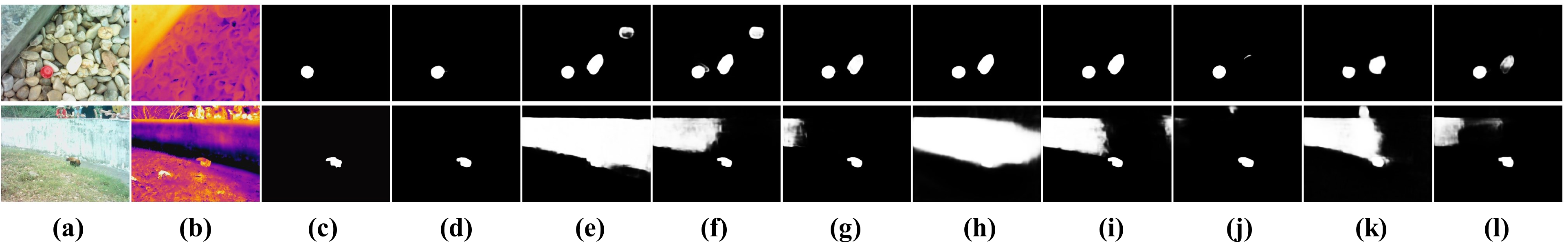}}
\caption{Visualization results of ablation studies. (a) RGB images. (b) Thermal images. (c) GT. (d) Full model. (e) Only RGB. (f) w/o GIE. (g) w/o SCP. (h) SCP-\textgreater{}Concat. (i) w/o LC. (j) w/o Localization. (k) w/o Complementation. (l) direct addition.
 }
\label{fig7}
\end{figure*}

To verify the effectiveness of the proposed GIE module, we conduct an experiment that removes the global illuminance score $\alpha$ from the full model, denotes as `w/o GIE'. 
As shown in the third row of Table \ref{tab2}, we can see that After removing the GIE module, the performance on all three datasets degrades. For example, the maximum F-measure is reduced from 0.895 to 0.880 on the VT5000 dataset.
As shown in Fig. \ref{fig7}, the thermal image contains some non-salient but hot objects, such as the wall in the second image, 
Therefore, without the adjustment of the GIE module, the model is affected by noisy objects in this scene, as shown in the sixth column of Fig. \ref{fig7}.
From the above two aspects, it can be seen that our proposed GIE module can better regulate the role of RGB features and thermal features, and can better utilize the features of the two modalities to achieve better performance.


\begin{table}[!t]
\renewcommand\arraystretch{1.5}
\caption{Ablation studies of key components in our TNet.}
\resizebox{\linewidth}{!}{
\begin{tabular}{c|c|cc|cc|cc}
\hline
\multirow{2}{*}{Type} & \multirow{2}{*}{Model}        & \multicolumn{2}{c|}{VT5000}                          & \multicolumn{2}{c|}{VT1000}                         & \multicolumn{2}{c}{VT821}                           \\ \cline{3-8} 
                      &                               & \multicolumn{1}{c|}{$F_{\beta}^m $}           & $MAE$            & \multicolumn{1}{c|}{$F_{\beta}^m $}           & $MAE$           & \multicolumn{1}{c|}{$F_{\beta}^m $}           & $MAE$            \\ \hline
\textbf{TNet}         & \textbf{Full model}           & \multicolumn{1}{c|}{\textbf{0.895}} & \textbf{0.033} & \multicolumn{1}{c|}{\textbf{0.937}} & \textbf{0.021} & \multicolumn{1}{c|}{\textbf{0.904}} & \textbf{0.030} \\ \hline
-                   & Only RGB                       & \multicolumn{1}{c|}{0.866}          & 0.043          & \multicolumn{1}{c|}{0.918}          & 0.029         & \multicolumn{1}{c|}{0.851}          & 0.051          \\ \hline
GIE                   & w/o GIE                       & \multicolumn{1}{c|}{0.880}          & 0.038          & \multicolumn{1}{c|}{0.931}          & 0.027         & \multicolumn{1}{c|}{0.876}          & 0.042          \\ \hline
\multirow{2}{*}{SCP}  & w/o SCP                       & \multicolumn{1}{c|}{0.887}          & 0.035          & \multicolumn{1}{c|}{0.930}          & 0.025         & \multicolumn{1}{c|}{0.878}          & 0.043          \\ \cline{2-8} 
                      & SCP-\textgreater{}Concat      & \multicolumn{1}{c|}{0.882}          & 0.038          & \multicolumn{1}{c|}{0.932}          & 0.024         & \multicolumn{1}{c|}{0.881}          & 0.039          \\ \hline
\multirow{4}{*}{LC}   & w/o LC                        & \multicolumn{1}{c|}{0.882}          & 0.038          & \multicolumn{1}{c|}{0.932}          & 0.024          & \multicolumn{1}{c|}{0.880}          & 0.042          \\ \cline{2-8} 
                      & w/o Localization              & \multicolumn{1}{c|}{0.885}          & 0.036          & \multicolumn{1}{c|}{0.934}          & 0.022         & \multicolumn{1}{c|}{0.882}          & 0.037          \\ \cline{2-8} 
                      & w/o Complementation                & \multicolumn{1}{c|}{0.873}          & 0.038          & \multicolumn{1}{c|}{0.924}          & 0.025         & \multicolumn{1}{c|}{0.865}          & 0.044          \\ \cline{2-8}
                       & direct addition & \multicolumn{1}{c|}{0.882}          & 0.037          & \multicolumn{1}{c|}{0.929}          & 0.025         & \multicolumn{1}{c|}{0.874}          & 0.043           \\ \hline
\end{tabular}}
\label{tab2}
\end{table}

\subsubsection{Effectiveness of SCP module}
First, we verify the effect of removing the SCP module, denoted as `w/o SCP', and the results are shown in the fourth row of Table \ref{tab2}. By comparing the first and fourth rows in Table \ref{tab2}, the performance drop illustrates the need for the SCP module. The main reason for the necessity of the module is that the module can use the highest layer of RGB branch to encode the semantic attribute on each layer of the thermal features, making the thermal image more suitable for the SOD task. The lack of this module will lead to unsatisfactory detection due to the thermal features, which can be qualitatively seen from the seventh column of Fig. \ref{fig7}. 
At the same time, in order to verify the design of the SCP module, we directly replace the SCP module with concatenation between the semantic mask and thermal features, denoted as `SCP-\textgreater{}Concat'.
As shown in the fifth row of Table \ref{tab2}, this semantic guidance approach does not even outperform the performance of removing the SCP module, which indirectly illustrates the effectiveness of our SCP module design.


\subsubsection{Effectiveness of LC module}
The proposed LC module implements RGB-dominated cross-modality feature decoding, we firstly verify the effect of the entire LC module, doneted as `w/o LC'. Specifically, the encoding stage is not changed, the localization guidance stage in the LC module is removed, and the encoding features of RGB and thermal features are directly added as the final skip connection result in the skip-connection complementation stage. As shown in the sixth row of Table \ref{tab2} and ninth column of Fig. \ref{fig7}, it can be seen that removing the LC module leads to obvious performance degradation, such as the maximum F-measure drops from 0.895 to 0.882 on the VT5000 dataset, and the ambient noise is detected.

Then, we evaluate the roles of two stages in the LC module by removing them separately in the full model, where `w/o Localization' means removing the localization guidance stage and `w/o Complementation' means removing the skip-connection complementation stage.
As reported in Table \ref{tab2}, the performance of the model will degrade after removing any stage. For example, on the VT5000 dataset, the performance of maximum F-measure is decreased from 0.895 to 0.885 after removing the location stage, and to 0.873 after removing the complementation stage.
From the visual examples shown in Fig. \ref{fig7}, we can see that the broken LC model will result in incomplete detection or introduce additional noise. The fundamental reason is that our proposed LC module makes good use of the valuable information provided by the thermal modality to a large extent. The thermal modality is not only to assist the localization and suppress noise in the localization guidance stage, but also to achieve adaptive complementation and improve the detection integrity in the skip-connection complementation stage.

In order to demonstrate the design of adaptive skip connections in the skip-connection complementation stage of the LC module, we add an ablation study by replacing it with the direct skip connections, that is, replacing the global illuminance score based weighted fusion with the direct addition. From the last row of Table \ref{tab2}, we can see that with the help of the proposed adaptive skip connections, the performance is improved compared with the direct addition manner. For example, on the VT5000 dataset, compared with using direct skip connection (\ie, direct addition), the F-measure score is improved from 0.882 to 0.895 with a percentage gain of 1.5\%, and the MAE is changed from 0.037 to 0.033 with a percentage gain of 10.8\%. The visualization results shown in the last column of Fig. \ref{fig7} also show that our adaptive skip connections can better suppress irrelevant interference. 

\subsection{Failure Cases and Future Work}
\begin{figure}[!t]
\centering
\centerline{\includegraphics[width=1\linewidth]{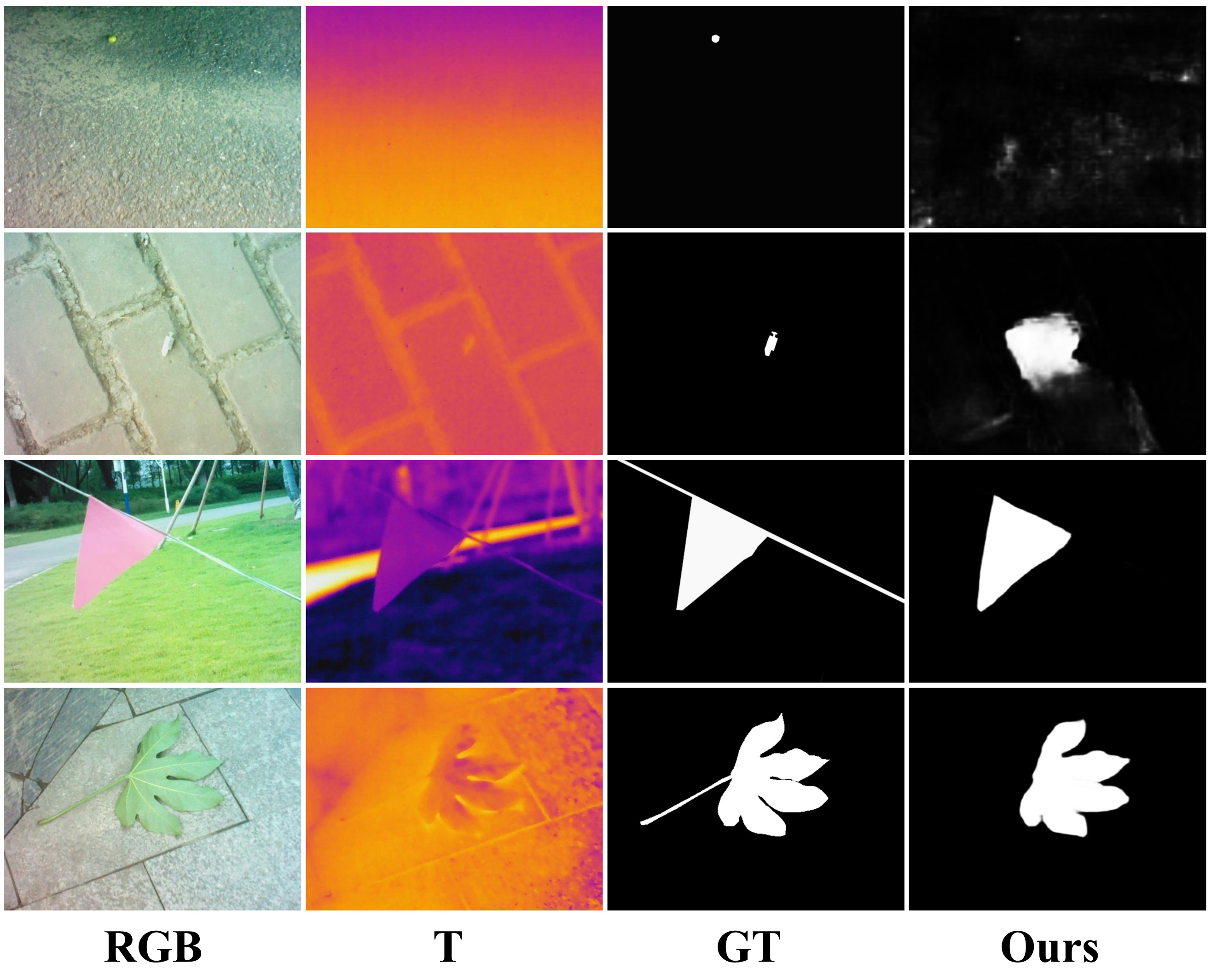}}
\caption{Failure cases of the proposed TNet.}
\label{fig_fcfw}
\end{figure}

In this paper, we rethink the connotation and role of thermal modality, thereby proposing a TNet to solve the RGB-T SOD task. On the one hand, we introduce a global illumination estimation module to regulate the role played by the two modalities. On the other hand, we design different interaction mechanisms (\ie, SCP module and LC module) in the encoding and decoding stages to achieve deep, comprehensive, and differentiated information interaction. Although our method shows competitive performance, it still falls short in some difficult scenarios, as shown in Fig. \ref{fig_fcfw}. For example, the first two rows are scenes in which salient objects are small and indistinguishable from the surrounding environment in RGB image and thermal image, inducing our proposed TNet cannot detect the salient objects correctly. Besides, in the third and fourth rows, the slender objects are difficult to be completely detected by our method because of their elongated structures and low contrast to the background. For such problems, we can resort to some stronger feature extractors (such as swin Transformer \cite{liu2021Swin}) and reasoning mechanisms (such as GNN \cite{crm/tgrs22/RRNet}) in the future. In addition, the existing public datasets for RGB-T SOD task (such as VT821, VT1000, and VT5000 datasets) do not fully consider the practical value of thermal images, where most of the scenes are daytime and the number of nighttime scenes is very limited. This is obviously inappropriate, therefore, it is necessary to build a large-scale RGB-T SOD dataset that includes more low-light scenes.

\section{Conclusion} \label{sec5}
In this paper, we propose a network that can adaptively adjust the roles of two modalities according to the brightness of the scene for RGB-T SOD task, called TNet. We introduce the GIE module that can estimate a global illuminance score of an image to regulate the cross-modality interaction. In the encoding stage, we semantically supplement the thermal features by using the semantic mask obtained from the RGB top layer features, so that the thermal features can be better adapted to the task of SOD. In the decoding stage, we utilize thermal features to provide better localization information and skip-connection supplementary information for RGB decoding features, thereby improving detection accuracy and completeness. Extensive experimental results on three datasets reveal the superiority of our model in handling the RGB-T SOD task.

\par
\ifCLASSOPTIONcaptionsoff
  \newpage
\fi
{
\bibliographystyle{IEEEtran}
\bibliography{ref}
}

\end{document}